\documentclass[a4paper,fleqn]{cas-dc}

\usepackage[numbers]{natbib}
\usepackage{amsmath,amssymb,amsfonts}
\usepackage{graphicx}
\usepackage{booktabs}
\usepackage{multirow}
\usepackage{url}
\usepackage{subfigure}
\usepackage{epstopdf}
\usepackage{makecell}
\usepackage{xcolor}
\usepackage{caption} 
\shortauthors{Chen et al.}

\title[mode=title]{Transformer-Guided Content-Adaptive Graph Learning for
Hyperspectral Unmixing}

\begin{document}
\RenewDocumentCommand{\printorcid}{}{}
\let\WriteBookmarks\relax
\def\floatpagepagefraction{1}
\def\textpagefraction{.001}

\shorttitle{Transformer-Guided Content-Adaptive Graph Learning for HU}

\tnotemark[1]
\tnotetext[1]{This work was supported in part by the Shanghai Natural Science Foundation of China under Grant 24ZR1425700, in part by the National Natural Science Foundation of China under Grant 12371306, and in part by the Project of the State Administration of Foreign Experts under Grant H20240974.}

\author[1]{Hui Chen}
\ead{chenhui@shiep.edu.cn}

\author[1]{Liangyu Liu}
\ead{liangyuliu@mail.shiep.edu.cn}

\author[2]{Xianchao Xiu}
\ead{xcxiu@shu.edu.cn}

\author[3]{Wanquan Liu}
\ead{liuwq63@mail.sysu.edu.cn}

\affiliation[1]{organization={School of Automation Engineering, Shanghai University of Electric Power},
                city={Shanghai},
                postcode={200090},
                country={China}}

\affiliation[2]{organization={School of Mechatronic Engineering and Automation, Shanghai University},
                city={Shanghai},
                postcode={200444},
                country={China}}

\affiliation[3]{organization={School of Intelligent Systems Engineering, Sun Yat-sen University},
                city={Guangzhou},
                postcode={510275},
                country={China}}

\cortext[cor1]{Corresponding author.}

\begin{abstract}
Hyperspectral unmixing (HU) aims to decompose each mixed pixel in remote sensing images into a set of endmembers and their corresponding abundances. Despite significant progress in this field using deep learning, most methods fail to simultaneously characterize global dependencies and local consistency, making it difficult to preserve both long-range interactions and boundary details. This paper proposes a novel transformer-guided content-adaptive graph unmixing framework (T-CAGU), which overcomes these challenges by employing a transformer to capture global dependencies and introducing a content-adaptive graph neural network to enhance local relationships. Unlike previous work, T-CAGU integrates multiple propagation orders to dynamically learn the graph structure, ensuring robustness against noise. Furthermore, T-CAGU leverages a graph residual mechanism to preserve global information and stabilize training. Experimental results demonstrate its superiority over state-of-the-art methods. Our code is available at \url{https://github.com/xianchaoxiu/T-CAGU}.
\end{abstract}


\maketitle
\section{Introduction}
Hyperspectral remote sensing imagery, capable of capturing rich and detailed spectral features, has found widespread applications in environmental monitoring \cite{guo2024saan}, geological exploration \cite{marques2024lithological}, target detection \cite{yang2024variational}, and agricultural assessment \cite{goswami2025exploring}. From a pattern recognition perspective, hyperspectral imagery has also been extensively studied for classification and target detection \cite{li2016survey}. However, due to the limited spatial resolution of imaging instruments, a single pixel often contains multiple components. To address this mixed pixel problem, hyperspectral unmixing (HU) techniques are employed to extract more detailed and accurate land cover information \cite{sun2024generative}.

According to \cite{bioucas2012hyperspectral}, traditional HU methods can be roughly divided into three categories: regression, geometric, and statistical. Among them, regression methods represent each mixed pixel as a linear combination of a limited set of pure spectral signatures from the spectral library \cite{huang2025hyperspectral}, while geometric methods typically exploit the data simplex or its positive cone, such as vertex component analysis (VCA) \cite{nascimento2005vertex}. From a statistical perspective, HU can be regarded as an inference problem, and non-negative matrix factorization (NMF) \cite{feng2022hyperspectral} has been highly recognized over the past few decades.

With the rapid development of deep learning and artificial intelligence, the applications to HU have also expanded rapidly. Ghosh et al. \cite{ghosh2022hyperspectral} proposed a transformer-based deep unmixing framework, called DeepTrans, which leverages the transformer's ability to capture global feature dependencies. However, this framework highly relies on initialization strategies such as VCA. To this end, Tao et al. \cite{tao2024abundance} constructed an abundance-guided attention spectral and spatial attention network (A2SAN) by learning endmembers and abundances in an end-to-end manner. Furthermore, Tao et al. \cite{tao2024new} developed a dual-feature fusion network (DFFN) to mitigate the randomness introduced by the autoencoder initialization through enhancing spectral/spatial similarity and modularity estimation, rather than directly replacing the initialization. Recently, Gao et al. \cite{gao2025ssaf} presented an unsupervised framework called spatial–spectral adaptive fusion network (SSAF-Net) that integrates scale perturbation models to address endmember variability and exhibits excellent unmixing performance.

\begin{figure*}[t]  
    \centering
    \includegraphics[scale=0.43]{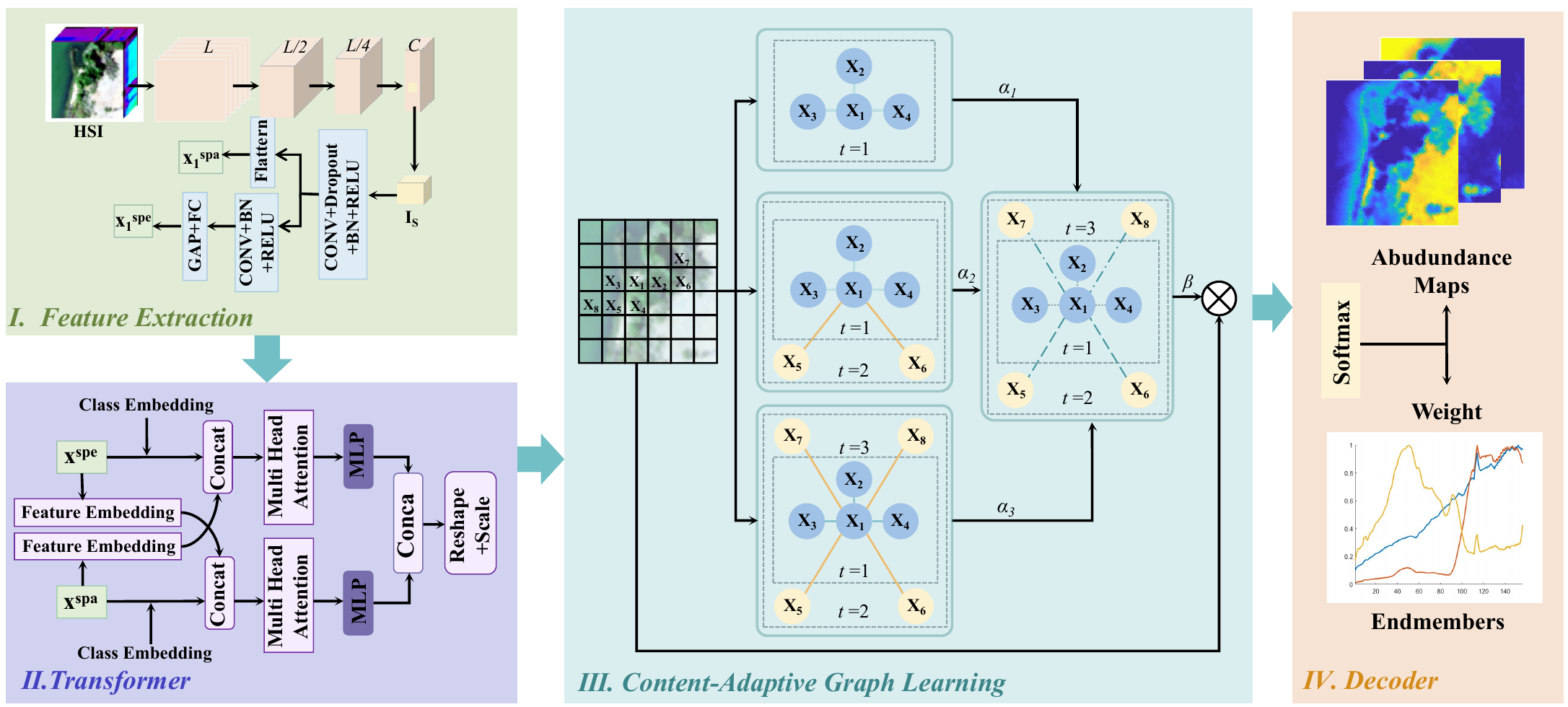}
    \vspace{-0.2cm}
     \caption{
\textit{Flowchart of our proposed T-CAGU. The original HSI is processed by a convolutional network to extract compact spectral–spatial features, which are then enhanced through a transformer to capture long-range  dependencies. A content-adaptive graph models local patches as nodes, dynamically updating relationships to learn correlations. In this process, different line styles indicate the learned edge strengths. Finally, a decoder reconstructs abundance maps with the softmax normalization while estimating endmembers.}}
    \label{Flowchart}
\end{figure*}

Note that graph learning has been used to capture hyperspectral information and encode spatial-spectral relationships. Jin et al. \cite{jin2023graph} incorporated the graph attention convolution into an autoencoder, enabling the use of both long- and short-range spatial information in HSIs for unsupervised nonlinear HU. Chen et al.\cite{chen2025adaptive} proposed an adaptive multi-order graph-regularized NMF that learns weights for multiple orders of graph propagation. Although these methods can capture local structure, they construct neighborhood graphs from spectral similarity and cannot characterize global dependencies.

Motivated by the above observations, this letter introduces a simple yet effective HU framework called transformer-guided content-adaptive graph unmixing (T-CAGU), which simultaneously captures global dependencies and local consistency. The main contributions are as follows.
\begin{enumerate}
\item We propose a novel unmixing framework that combines the advantages of the transformer for global dependencies and graph learning in maintaining local consistency, thereby achieving complementary strengths of both paradigms.
\item We design a content-adaptive graph construction module that dynamically adjusts the weights of multiple propagation orders during training, enabling multi-order information fusion and adaptive graph structure learning. This module is further incorporated as a graph propagation residual to preserve global information while enhancing local consistency and boundary details.
\item T-CAGU adopts a guided fusion mechanism in which the global representations from the transformer explicitly guide the adaptive learning of the graph structure. This hierarchical and dynamic fusion design has been rarely explored in existing studies and demonstrates superior performance in our experiments.
\end{enumerate}


\section{The Proposed Method}\label{sec:notations}

Assume that the spectral response of a pixel is a mixture of multiple endmembers, 
where the proportion of each endmember corresponds to its abundance. 
The linear mixing model (LMM) can be described as 
\begin{equation}
\mathbf{S} = \mathbf{E}\mathbf{M},
\end{equation}
where $\mathbf{S} = [\mathbf{s}_1, \mathbf{s}_2, \ldots, \mathbf{s}_N] \in \mathbb{R}^{L \times N}$ 
is the spectral matrix of the HSI, consisting of $N$ pixel spectral vectors with $L$ spectral bands, 
and $N = H \times W$, of which $H$ and $W$ represent the height and width, respectively. 
$\mathbf{E} \in \mathbb{R}^{L \times P}$ is the endmember matrix, where $P$ denotes the number of endmembers. 
$\mathbf{M} \in \mathbb{R}^{P \times N}$ is the abundance matrix, which indicates the proportion 
of each endmember in each pixel.

In order to estimate abundance $\mathbf{M}$, an autoencoder-based network is often employed. 
Specifically, the encoder $f^{en}(\cdot)$ maps the high-dimensional input $\mathbf{S}$ 
to a low-dimensional representation space, given by
\begin{equation}
\mathbf{M} = f^{en}(\mathbf{S}),
\end{equation}
where the latent representation $\mathbf{M}$ corresponds to the abundance matrix, 
which is consistent with the unmixing process. 
The decoder $f^{de}(\cdot)$ then reconstructs the HSI from $\mathbf{M}$ as 
\begin{equation}
\hat{\mathbf{S}} = f^{de}(\mathbf{M}) = \mathbf{E}\mathbf{M},
\end{equation}
where $\hat{\mathbf{S}}$ represents the reconstructed HSI, 
and the decoder weight matrix $\mathbf{E}$ is interpreted as the endmember matrix.

As shown in Fig. \ref{Flowchart}, our proposed T-CAGU consists of four parts, which will be explained in detail below.

\subsection{Feature Extraction}

A spectral compression strategy is employed to reduce the number of channels while extracting discriminative features as input to the transformer. This process can be formulated as
\begin{equation}
    \mathbf{I}' = f_{\theta}(\mathbf{I}), 
    \quad f_{\theta}: \mathbb{R}^{L \times H \times W} \rightarrow \mathbb{R}^{C \times H \times W},
\end{equation}
where $\theta = \{\mathbf{W}_1, \mathbf{W}_2, \mathbf{W}_3, \mathbf{V}_1, \mathbf{V}_2, \mathbf{V}_3\}$ with $\mathbf{W}_i$ and $\mathbf{V}_i$ denoting the weight matrix and bias of the $i$-th convolutional layer, respectively. $f_{\theta}$ represents the mapping function composed of three stacked convolutional layers. The input HSI $\mathbf{I} \in \mathbb{R}^{L \times H \times W}$ is progressively compressed through three convolutional operations, where the number of spectral channels is reduced from $L \rightarrow \tfrac{L}{2} \rightarrow \tfrac{L}{4} \rightarrow C$, resulting in $\mathbf{I}' \in \mathbb{R}^{C \times H \times W}$. Here, $C$ is a hyperparameter.

After obtaining the feature map $\mathbf{I}'$, a block of size $m \times m$ is extracted. Along the spectral dimension, a $1 \times 1$ convolution, pooling, and a fully connected layer are applied to extract spectral structural information, yielding the spectral feature vector 
$\mathbf{x}_1^{spe} \in \mathbb{R}^{1 \times D}$. 
Along the spatial dimension, a $3 \times 3$ convolution is employed to extract spatial structural information, resulting in the spatial feature vector 
$\mathbf{x}_1^{spa} \in \mathbb{R}^{1 \times S}$.

By traversing all patches, it is easy to obtain the spectral feature sequence, i.e.,
\begin{equation}
\mathbf{X}^{spe} = \left\{\mathbf{x}^{spe}_1, \mathbf{x}^{spe}_2, \dots, \mathbf{x}^{spe}_n \right\},
\end{equation}
and the spatial feature sequence
\begin{equation}
\mathbf{X}^{spa} = \left\{\mathbf{x}^{spa}_1, \mathbf{x}^{spa}_2, \dots, \mathbf{x}^{spa}_m \right\},
\end{equation}
where $n$ and $m$ are the number of spectral tokens and the number of spatial tokens extracted from all patches, respectively.

\subsection{Transformer}

To introduce global information from the counterpart branch at the attention stage, it enables the two branches to share each other’s global information, thereby promoting spectral--spatial feature alignment. The class token from the spatial branch $\mathbf{cls}^{spa}$ is prepended to the spectral sequence, forming  
\begin{equation}
\tilde{\mathbf{X}}^{spe} = [\mathbf{cls}^{spa}, \mathbf{X}^{spe}],
\end{equation}
while the class token from the spectral branch $\mathbf{cls}^{spe}$ is prepended to the spatial sequence, yielding
\begin{equation}
\tilde{\mathbf{X}}^{spa} = [\mathbf{cls}^{spe}, \mathbf{X}^{spa}].
\end{equation}

The sequences $\tilde{\mathbf{X}}^{spe}$ and $\tilde{\mathbf{X}}^{spa}$ are then fed into the transformer self-attention module. 
Given an input sequence $\tilde{\mathbf{X}} \in \{\tilde{\mathbf{X}}^{spe}, \tilde{\mathbf{X}}^{spa}\}$, the query, key, and value matrices are computed in the form of
\begin{equation}
Q = \tilde{\mathbf{X}} W_Q, \quad 
K = \tilde{\mathbf{X}} W_K, \quad 
V = \tilde{\mathbf{X}} W_V,
\end{equation}
where $\mathbf{W}_Q, \mathbf{W}_K, \mathbf{W}_V$ are the learnable projection matrices. 
The scaled dot-product attention is then given by  
\begin{equation}
\tilde{\mathbf{X}}' = \text{Attention}(\mathbf{Q}, \mathbf{K}, \mathbf{V}) 
= \text{Softmax}\!\left(\frac{\mathbf{Q}\mathbf{K}^\top}{\sqrt{d_k}}\right)\mathbf{V},
\end{equation}
where $d_k$ is the dimension size of the key vectors, and $\tilde{\mathbf{X'}} \in \{\tilde{\mathbf{X'}}^{spe}, \tilde{\mathbf{X'}}^{spa}\}$.

By combining the outputs via residual addition and multilayer perceptron (MLP) layers, it produces the joint spectral-spatial representation, given by
\begin{equation}
\mathbf{I}_E = \text{Concat}[\, \text{MLP}(\tilde{\mathbf{X'}}^{spe}), \, \text{MLP}(\tilde{\mathbf{X'}}^{spa}) \,].
\end{equation}

Subsequently, through the upsampling operation, the feature map is restored to the same spatial size as the input image, which can be formulated as  
\begin{equation}
\mathbf{I}'' = U(\mathbf{I}_E),
\end{equation}
where $U(\cdot)$ denotes the reshaping operation that reshapes 
$\mathbf{I}_E \in \mathbb{R}^{B \times dim}$ into 
$\mathbf{I}'' \in \mathbb{R}^{B \times H \times W}$, and $dim$ denotes the feature embedding dimension after spectral band mapping.

\subsection{Content-Adaptive Graph Learning}
After obtaining the feature map $\mathbf{I}''$ from the transformer, 
a content-adaptive graph learning module is introduced to refine the abundances 
in a residual manner. Given data $\mathbf{I}''$, each pixel is treated as a graph node, and the vertex set 
$\mathcal{V}$ is of size $N = H \times W$. For each node $i$, the spatial neighborhood $\mathcal{N}_r(i)$ 
is defined as all pixels within a window of radius $r$. The edge weight $D_{ij}$ is determined by spectral similarity and spatial proximity, i.e.,
\begin{equation}
D_{ij}
= \exp\!\Bigg(-\frac{\lVert \mathbf{f}_i - \mathbf{f}_j \rVert_2^2}{\sigma_f}\Bigg)
\cdot
\exp\!\Bigg(-\frac{\lVert \mathbf{g}_i - \mathbf{g}_j \rVert_2^2}{\sigma_g^2}\Bigg),
\label{eq:content_adaptive_weight_euclidean}
\end{equation}
where $\sigma_f,\sigma_g > 0$ are the scaling parameters. $\mathbf{f}_i$ is the feature vector of pixel $i$ generated by the transformer and $\mathbf{g}_i$ denotes the spatial position of pixel $i$. With dynamic graph construction, the adjacency and edge weights are updated according to $\mathbf{f}_i$, thereby adapting during training toward reducing the unmixing error. Compared with a static graph constructed directly from raw spectra, which is susceptible to noise perturbations, the transformer features $\mathbf{f}_i$ incorporate contextual information, possess stronger global semantics, and are more robust. Therefore, the transformer provides global dependencies, while graph propagation based on the dynamic graph emphasizes local consistency. Together, they enforce a global-to-local consistency constraint.

The weighted adjacency matrix $\mathbf{A}$ is defined as 
\begin{equation}
A_{ij} =
\begin{cases}
D_{ij}, & j \in \mathcal{N}_r(i),\ j \neq i,\\
0,      & \text{otherwise}.
\end{cases}
\label{eq:A_ij}
\end{equation}

For stable propagation, self-loops are added and symmetric normalization is applied
\begin{equation}
\tilde{\mathbf{A}} = \mathbf{A} + \text{diag}(\mathbf{1}), \quad \mathbf{D}_{ii} = \sum_{j} \tilde{\mathbf{A}}_{ij}, \quad \hat{\mathbf{A}} = \mathbf{D}^{-\tfrac{1}{2}} \tilde{\mathbf{A}} \mathbf{D}^{-\tfrac{1}{2}},
\end{equation}
where $\text{diag}(\mathbf{1})$ denotes the identity matrix. Let $\mathbf{X} \in \mathbb{R}^{B \times N}$ be the matrix obtained by unfolding the feature map $\mathbf{I}''$.
A channel projection is first applied as
\begin{equation}
\mathbf{Z}^{(0)} = \mathbf{M}\mathbf{X},
\end{equation}
where $\mathbf{M} \in \mathbb{R}^{B \times B}$ is a learnable projection matrix. 
To enable continuous information diffusion over the graph, $K$ propagation steps are performed, obtaining
\begin{equation}
\mathbf{Z}^{(t)} = \mathbf{Z}^{(t-1)} \hat{\mathbf{A}}, \quad t = 1, 2, \ldots, K,
\end{equation}

To avoid over-smoothing and adaptively select receptive fields, 
learnable weights are introduced as
\begin{equation}
\mathbf{Y} = \sum_{t=1}^{K} \alpha_t \mathbf{Z}^{(t)},
\end{equation}
where $\boldsymbol{\alpha} = \{\alpha_1, \alpha_2, \ldots, \alpha_K\}$ denotes a set of learnable weights that are non-negative and sum to one. 
In this way, the network can adaptively select the most appropriate combination. Let
\begin{equation}
\mathbf{X}' = \mathbf{X} + \beta \mathbf{Y},
\end{equation}
where $\beta$ is a hyperparameter. To decode, $\mathbf{X}'\in\mathbb{R}^{B\times N}$ is folded back into a tensor of size $\mathbb{R}^{B\times H\times W}$.

\subsection{Decoder}
The decoder uses a four-layer \(1\times1\) convolutional trunk to reduce the channel dimension to the number of endmembers \(P\), sharing parameters between the reconstruction and abundance branches for joint learning. A \(3\times3\) convolution followed by a Softmax produces the abundance map \(\mathbf{M}\), ensuring non-negativity and the sum-to-one constraint. 
A subsequent normalization step improves numerical stability. Finally, \(\mathbf{M}\) is projected through a \(1\times1\) convolution to reconstruct the spectrum \(\hat{\mathbf{I}}\), whose weights correspond to the VCA-initialized endmember spectra, preserving their physical meaning.

To train the model, the loss is defined as 
\begin{equation}
L = L_{\text{MSE}}(\mathbf{I}, \hat{\mathbf{I}})  + L_{\text{SAD}}(\mathbf{I}, \hat{\mathbf{I}}),
\end{equation}
where 
\begin{equation}
L_{\text{MSE}}(\mathbf{I}, \hat{\mathbf{I}}) 
= \frac{1}{H \cdot W} \sum_{i=1}^{H} \sum_{j=1}^{W} 
\left\| \hat{\mathbf{I}}_{ij} - \mathbf{I}_{ij} \right\|_2^2,
\end{equation}
\begin{equation}
L_{\text{SAD}}(\mathbf{I}, \hat{\mathbf{I}}) 
= \frac{1}{P} \sum_{i=1}^{P} 
\arccos \left( 
\frac{ \langle \mathbf{I}_i, \hat{\mathbf{I}}_i \rangle}
{ \|\mathbf{I}_i\|_2 \, \|\hat{\mathbf{I}}_i\|_2}
\right),
\end{equation}
are the mean squared error (MSE) and spectral angle distance (SAD), respectively.

\section{Experimental Results}\label{Experimental Results}


This section conducts numerical comparisons of our proposed T-CAGU with benchmark HU methods, including DeepTrans \cite{ghosh2022hyperspectral}, A2SAN \cite{tao2024abundance}, DFNN  \cite{tao2024new}, and SSAF-Net \cite{gao2025ssaf}.

In the experiments, the endmembers extracted by the VCA algorithm are used to initialize the weights of the decoder. Besides,  SAD and root-mean-square error (RMSE) are selected as evaluation metrics to assess the unmixing performance, with the best results are in bold, and second-best results are underlined.

\subsection{Results on Simulated Datasets}\label{Results on Synthetic Dataset}

A simulated dataset is generated based on \cite{hendrix2011new}, consisting of $100 \times 100$ pixels and 221 spectral bands. In this study, a learning rate of 1e-3, a weight decay of 1e-5, and 200 epochs are used. In addition, the balancing parameter $\beta$ is set to 0.2.

It is concluded from Table \ref{tab:synthetic_sad_rmse} that, as the SNR increases, the SAD and RMSE values of all compared methods decrease. Overall, the proposed T-CAGU achieves the best or second-best performance across a wide range of noise levels, demonstrating its effectiveness and robustness. This is attributed to the transformer prior, which enables the network to implicitly fit more faithful hyperspectral and abundance maps, the learnable dynamic graph propagation, which suppresses high-frequency noise, and the residual injection, which preserves high-frequency boundary details.

\begin{table}[t]
\footnotesize
\renewcommand{\arraystretch}{1.2}
\caption{SAD and RMSE results on the simulated datasets under different noise levels. }
\label{tab:synthetic_sad_rmse}
\vspace{-0.05cm}
\centering
\setlength{\tabcolsep}{0.7mm}
\begin{tabular}{|c|c|c|c|c|c|c|}
\hline
SNR & Metric & DeepTrans & A2SAN & DFNN & SSAF-Net & T-CAGU \\
\hline\hline
\multirow{2}{*}{10}
 & SAD   & 0.0952 & \underline{0.0711} & 0.0912 & 0.1157 & \textbf{0.0692} \\
\cline{2-7}
 & RMSE  & 0.2358 & 0.1807 & \textbf{0.1462} & 0.2771 & \underline{0.1708} \\
\hline
\multirow{2}{*}{20}
 & SAD   & 0.0860 & \underline{0.0699} & 0.0892 & 0.0929 & \textbf{0.0671} \\
\cline{2-7}
 & RMSE  & 0.2047 & 0.1782 & \underline{0.1563} & 0.2146 & \textbf{0.1192} \\
\hline
\multirow{2}{*}{30}
 & SAD   & \underline{0.0616} & 0.0698 & 0.0704 & 0.0858 & \textbf{0.0584} \\
\cline{2-7}
 & RMSE  & 0.1994 & 0.1846 & \underline{0.1837} & 0.1918 & \textbf{0.1115} \\
\hline
\multirow{2}{*}{40}
 & SAD   & \underline{0.0496} & 0.0592 & 0.0739 & 0.0663 & \textbf{0.0092} \\
\cline{2-7}
 & RMSE  & 0.1711 & \underline{0.1380} & 0.1845 & 0.1871 & \textbf{0.1071} \\
\hline\hline
\multirow{2}{*}{Mean}
 & SAD   & 0.0731 & \underline{0.0675} & 0.0812 & 0.0902 & \textbf{0.0510} \\
\cline{2-7}
 & RMSE  & 0.2028 & 0.1704 & \underline{0.1677} & 0.2177 & \textbf{0.1272} \\
\hline
\end{tabular}
\end{table}

\subsection{Results on Real Datasets}\label{Results on Real-world Datasets}

\begin{figure}
    \centering
    \includegraphics[width=1\linewidth]{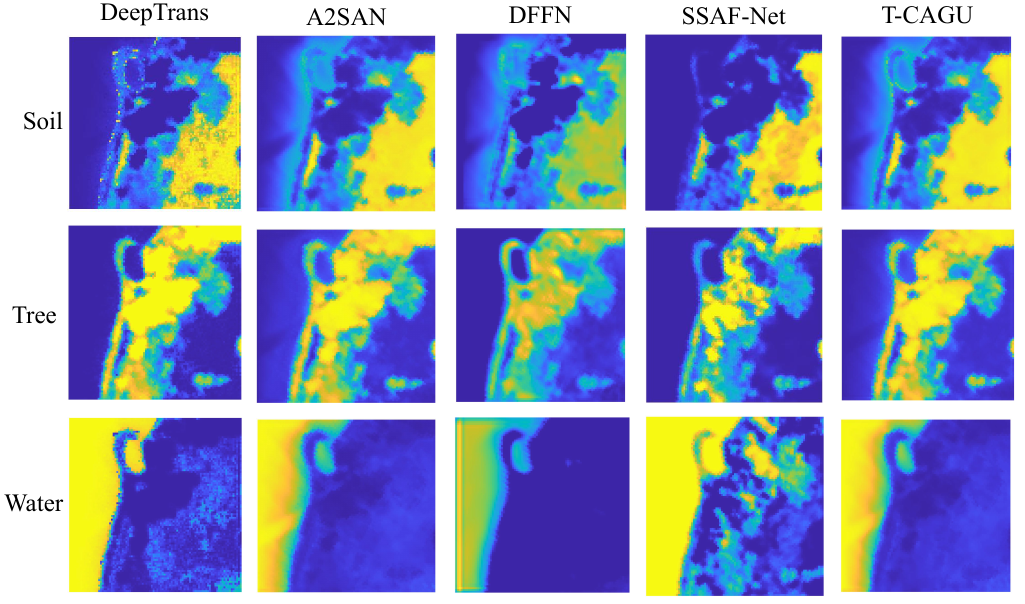} 
    \vspace{-0.6cm}
    \caption{Abundance map comparisons on the Samson dataset.
}
    \label{samsonrmse}
\end{figure}

In the Samson dataset\footnote{\url{http://www.escience.cn/people/feiyunZHU/Dataset_GT.html}}, it contains $952\times952$ pixels and 156 bands, spanning 401–889~nm. A $95\times95$ subimage with the upper-left corner at $(252,332)$ is analyzed. The scene comprises three endmembers, i.e., soil, tree, and water. For this dataset, the learning rate is 6e-4, the weight decay is 2e-5, the training epochs are 200, and the balance parameter $\beta$ is 0.2.
Table \ref{tab:samson_cmp} reports the SAD results for each endmember, and Fig. \ref{samsonrmse} visualizes the associated estimated abundance maps, where blue indicates low abundances and yellow indicates high abundances. It can be seen that the proposed T-CAGU consistently performs the best and produces clearer and better-separated abundance maps, which demonstrates the effectiveness of our proposed method.

\begin{table}[t]
\renewcommand{\arraystretch}{1.1}
\footnotesize
\caption{SAD comparisons on the Samson dataset. Best results are in bold,
and second-best results are underlined.}
\label{tab:samson_cmp}
\vspace{-0.15cm}
\centering
\setlength{\tabcolsep}{1mm}
\begin{tabular}{|c|c|c|c|c|c|}
\hline
Case & DeepTrans & A2SAN & DFNN & SSAF-Net & T-CAGU \\
\hline\hline
Soil  & \underline{0.0221} & 0.0686 & 0.0269 & 0.0253 & \textbf{0.0206} \\
\hline
Tree  & 0.0853 & 0.0422 & 0.0733 & \underline{0.0393} & \textbf{0.0308} \\
\hline
Water & \underline{0.0705} & 0.0740 & 0.1659 & 0.1334 & \textbf{0.0516} \\
\hline\hline
Mean  & \underline{0.0593} & 0.0616 & 0.0887 & 0.0660 & \textbf{0.0343} \\
\hline
\end{tabular}
\end{table}

\begin{table}[H]
\renewcommand{\arraystretch}{1.1}
\footnotesize
\caption{SAD comparisons on the Jasper dataset. Best results are in bold,
and second-best results are underlined.}
\label{tab:jasper_cmp}
\vspace{-0.15cm}
\centering
\setlength{\tabcolsep}{1mm}
\begin{tabular}{|c|c|c|c|c|c|}
\hline
Case & DeepTrans & A2SAN & DFNN & SSAF-Net & T-CAGU \\
\hline\hline
Tree  & 0.1062 & 0.0990 & 0.1526 & \underline{0.0891} & \textbf{0.0807} \\
\hline
Water & \textbf{0.0219} & 0.0914 & 0.1402 & 0.1192 & \underline{0.0284} \\
\hline
Soil  & \textbf{0.0837} & 0.2051 & 0.1081 & 0.1162 & \underline{0.1073} \\
\hline
Road  & 0.0932 & 0.0935 & 0.2783 & \underline{0.0379} & \textbf{0.0371} \\
\hline\hline
Mean  & \underline{0.0763} & 0.1223 & 0.1698 & 0.0906 & \textbf{0.0634} \\
\hline
\end{tabular}
\end{table}

\begin{figure}
    \centering
    \includegraphics[width=1\linewidth]{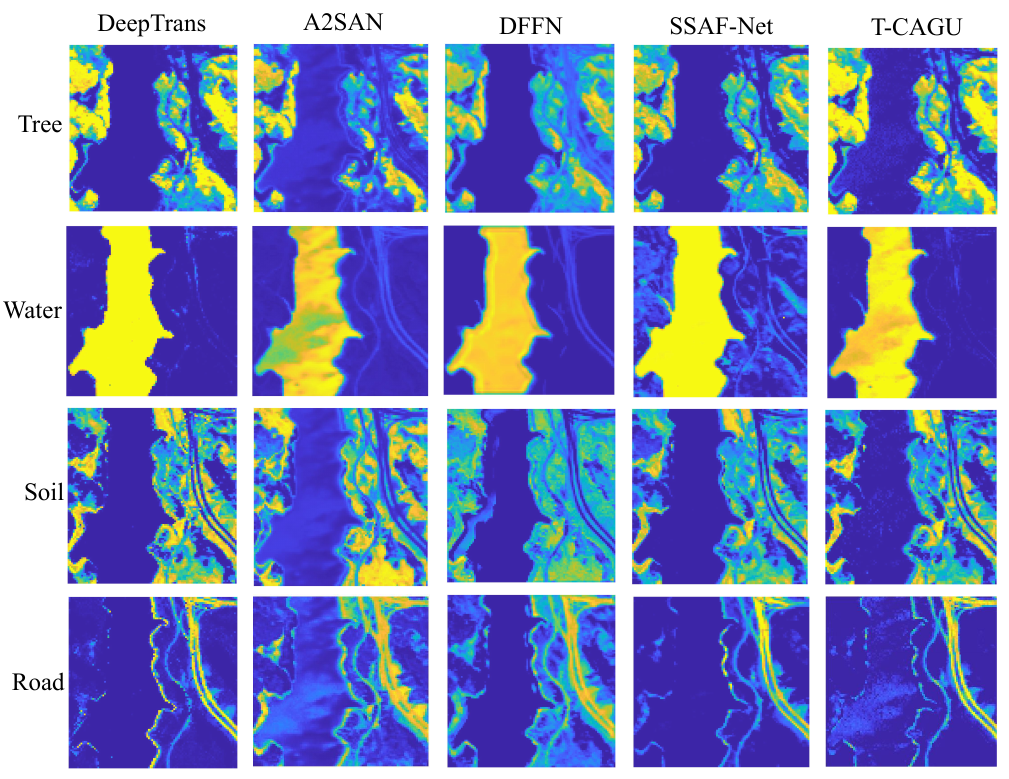} 
    \vspace{-0.6 cm}
    \caption{Abundance map comparisons on the Jasper dataset.
}
    \label{jasperrmse}
\end{figure}

In the Jasper Ridge dataset\footnote{\url{http://www.escience.cn/system/file?fileId=68574}}, it contains $512\times614$ pixels and 224 bands, spanning 380–2500~nm. A $100\times100$ subimage is analyzed. The scene comprises four endmembers, i.e., tree, water, soil, and road. Different from the above experiments, the learning rate is 8e-4 and the weight decay is 4e-5. The results in Table \ref{tab:jasper_cmp} confirm that the proposed T-CAGU also achieves the best mean SAD. The abundance maps in Fig. \ref{jasperrmse} further exhibit smoother and more realistic spatial distributions. 

\begin{figure}
  \centering
  \includegraphics[width=0.45\textwidth]{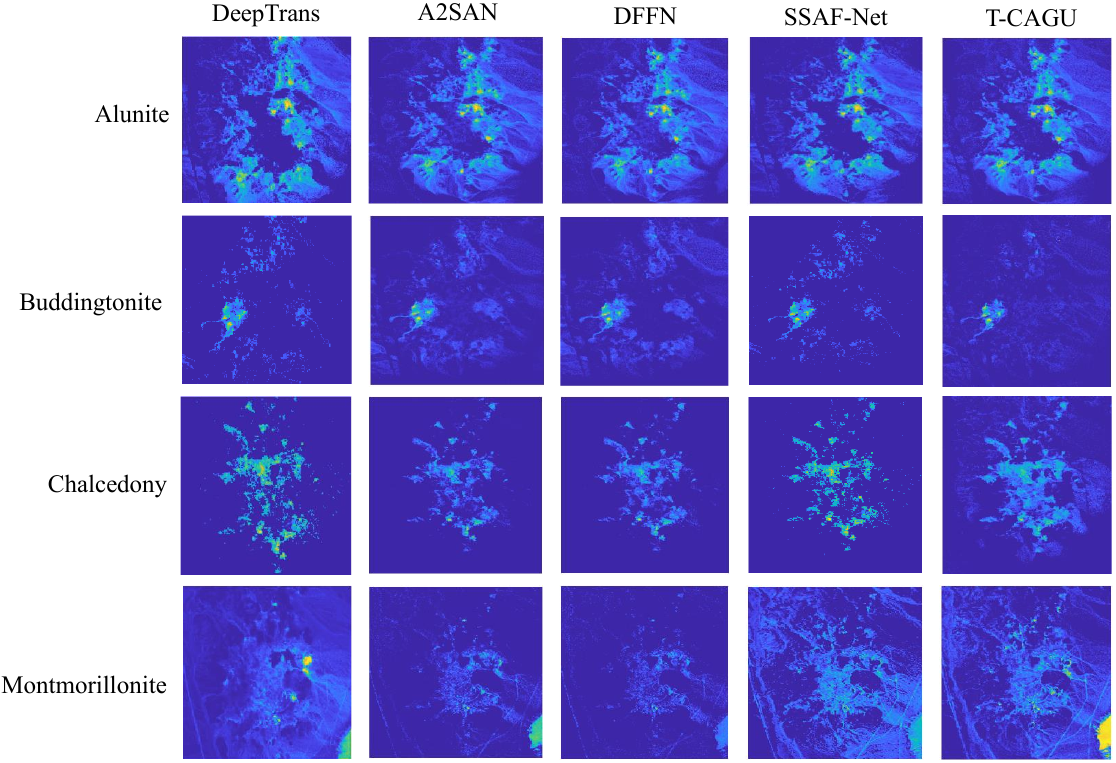}
  \vspace{-0.2cm}
  \caption{Abundance map comparisons on the Cuprite dataset.}
  \label{Abundancecuprite}
\end{figure}

This Cuprite dataset provides a mineralogical map\footnote{https://www.usgs.gov/media/images/aviris-scene-flown-over-cuprite-nevada}, enabling qualitative visual comparison. Specifically, we present estimated abundance maps of several representative minerals, including alunite, buddingtonite, chalcedony, and montmorillonite, and compare our results with some recent deep learning-based methods. As shown in Fig. \ref{Abundancecuprite}, our method produces abundance distributions that are more consistent with the known mineralogical spatial patterns, further demonstrating the effectiveness of the proposed method on real hyperspectral scenes.

\subsection{Ablation Studies}
To evaluate the contribution of the graph module, three cases are considered: I) without graph propagation; II) with a static grid graph; and III) with a dynamic graph.

\begin{figure}
	\setlength{\subfigcapskip}{-5pt}  
	\subfigure[SAD]{
		\includegraphics[scale=0.14]{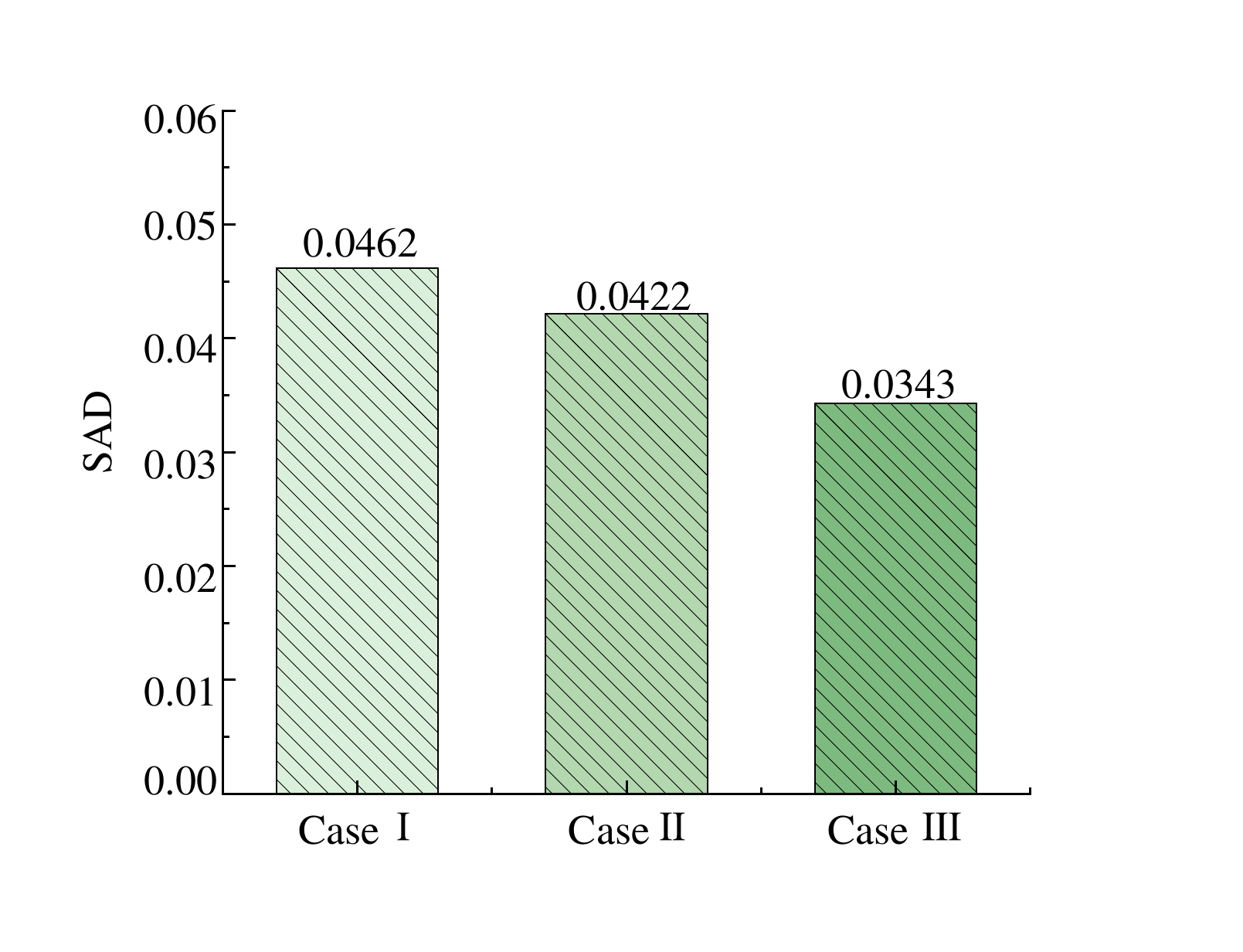} 
	}
	\subfigure[RMSE]{			
		\includegraphics[scale=0.14]{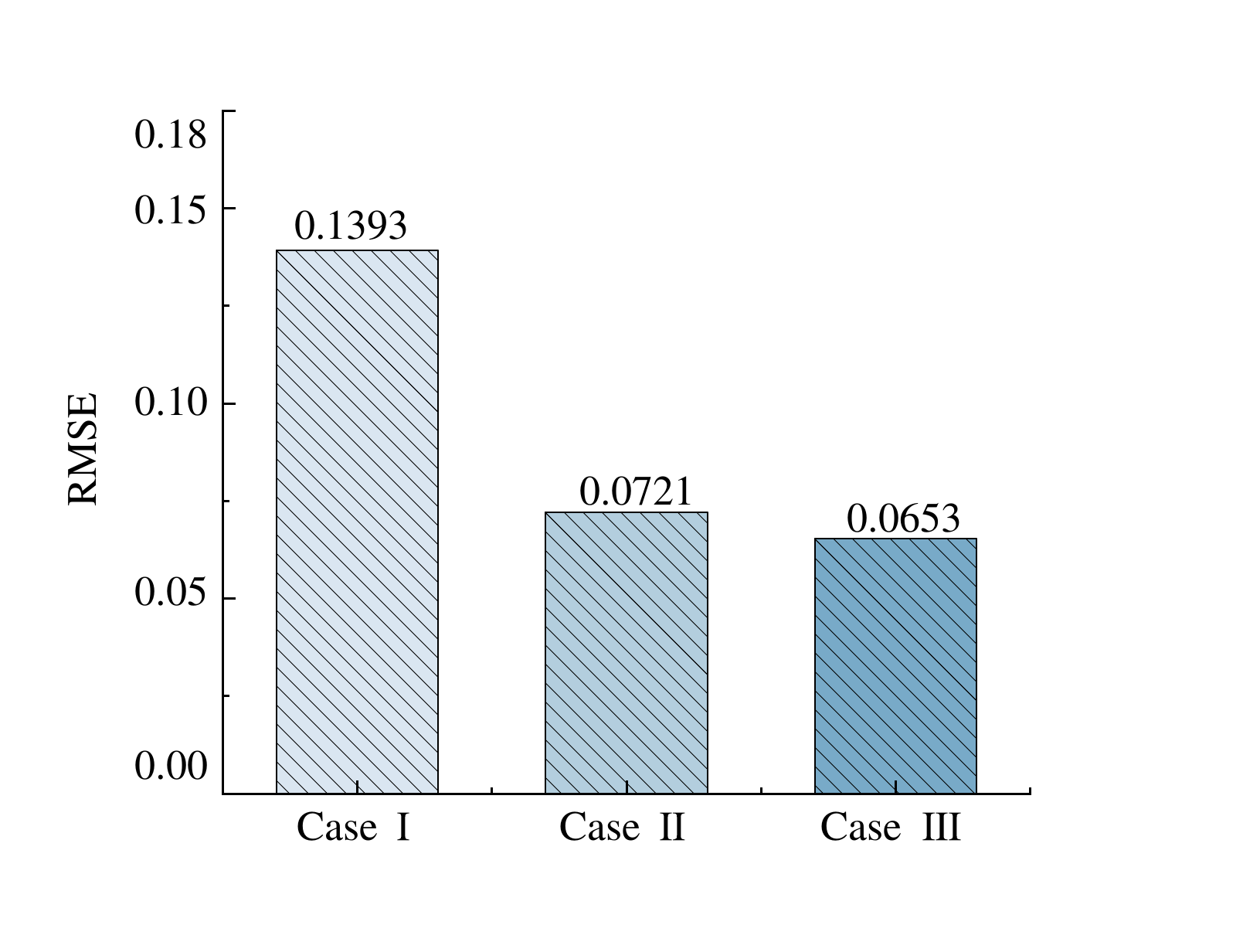}
	}
	\vspace{-0.2cm}
	\caption{Ablation studies of the graph module on the Samson dataset.}\label{fig:ablation1}
\end{figure}

As can be observed from Fig. \ref{fig:ablation1}, Case III achieves better unmixing results than Case I and Case II. Specifically, the static graph provides some improvements in local features. In contrast, the dynamic graph adaptively updates edge weights based on learned features, achieving more robust local refinement and sharper boundaries. This effectively complements the transformer's global modeling capabilities and achieves a better global-local balance. The ablation studies highlight the effectiveness of content-adaptive graph learning in improving local consistency and preserving boundaries.

\subsection{Discussion of Parameter $\beta$}	

\begin{figure}
    \setlength{\subfigcapskip}{-5pt} 
    \centering
    \subfigure[SAD]{
        \includegraphics[scale=0.23]{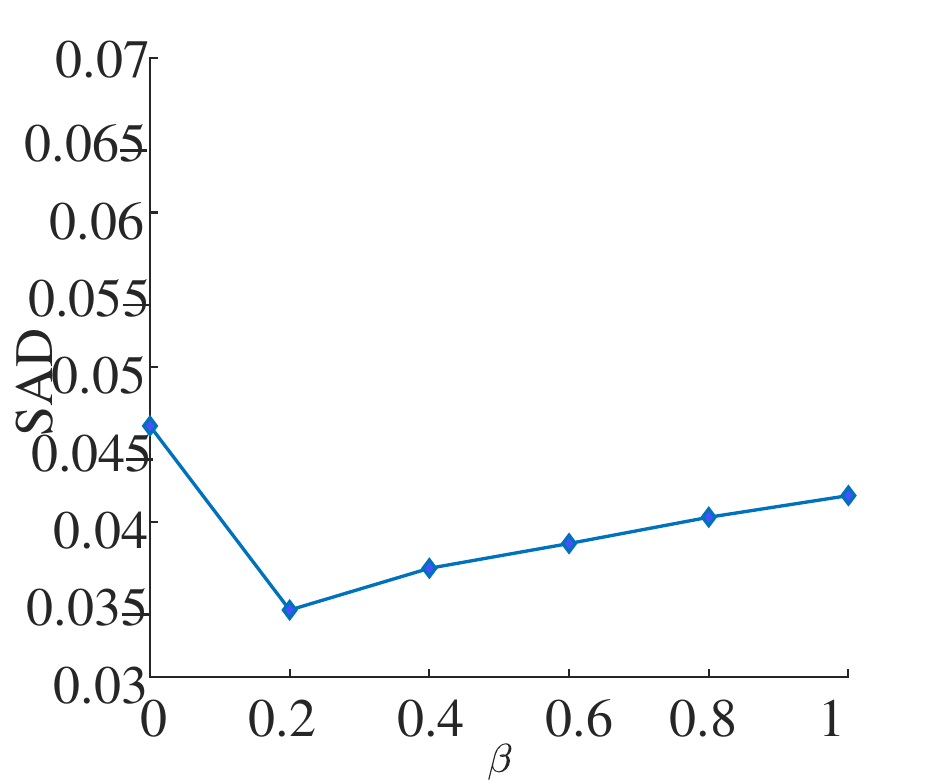}
    }
    \subfigure[RMSE]{
        \includegraphics[scale=0.23]{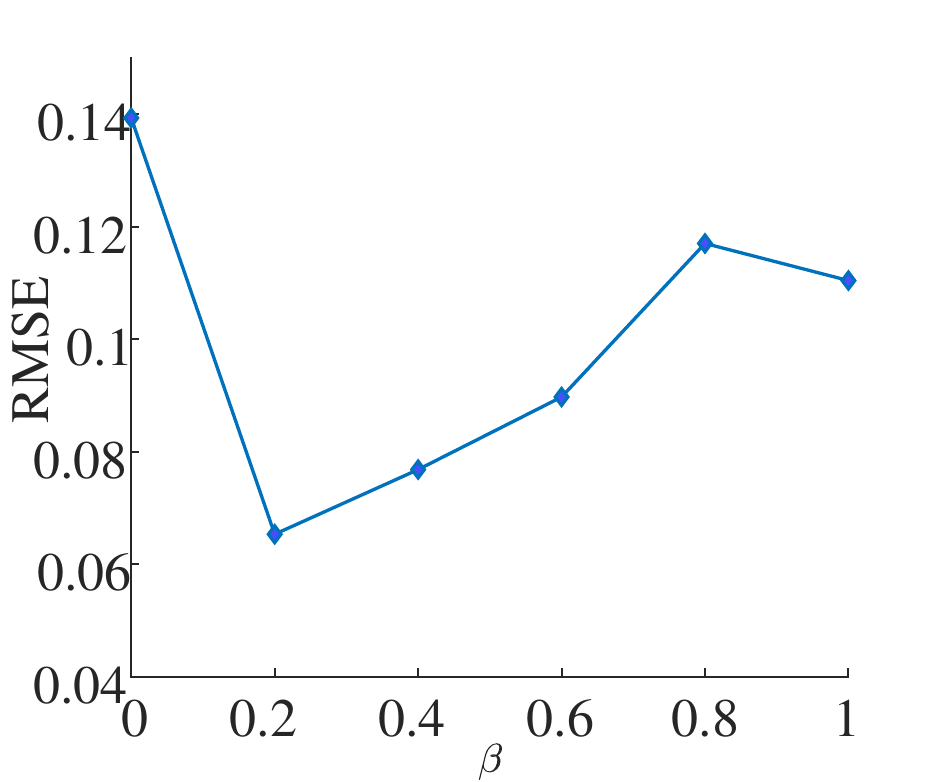}
    }
    \vspace{-0.2cm}
    \caption{Impact of $\beta$ on the Samson dataset.}
    \label{fig:ablation2}
\end{figure}

To discuss the impact of residual strength, a sensitivity study of $\beta$ is conducted on the Samson dataset. The parameter $\beta$ is varied within $\{0,0.2,\ldots,1\}$ while keeping all other settings fixed. As shown in Fig. \ref{fig:ablation2}, the residual injection strength plays a critical role in model performance. 
When $\beta = 0$, the absence of residual connections weakens local consistency and leads to performance degradation. 
When $\beta$ approaches $1$, over-smoothing occurs and boundary details are lost. The best unmixing performance is achieved when $\beta = 0.2$.

\subsection{Impact of Order $K$}
\begin{figure}
	\setlength{\subfigcapskip}{-2pt}  
	\subfigure[ ]{
		\includegraphics[scale=0.14]{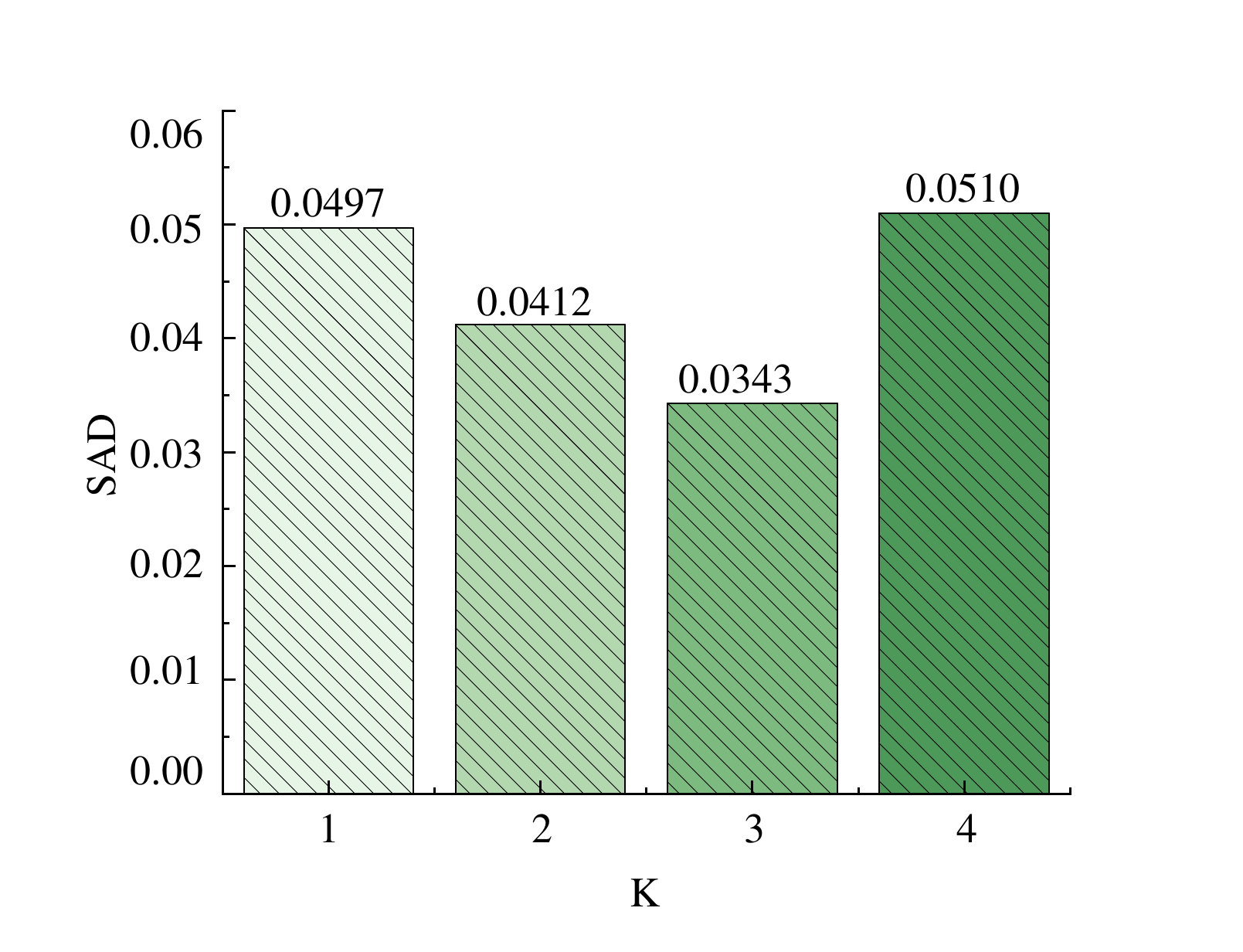}
	}
	\subfigure[ ]{			
		\includegraphics[scale=0.14]{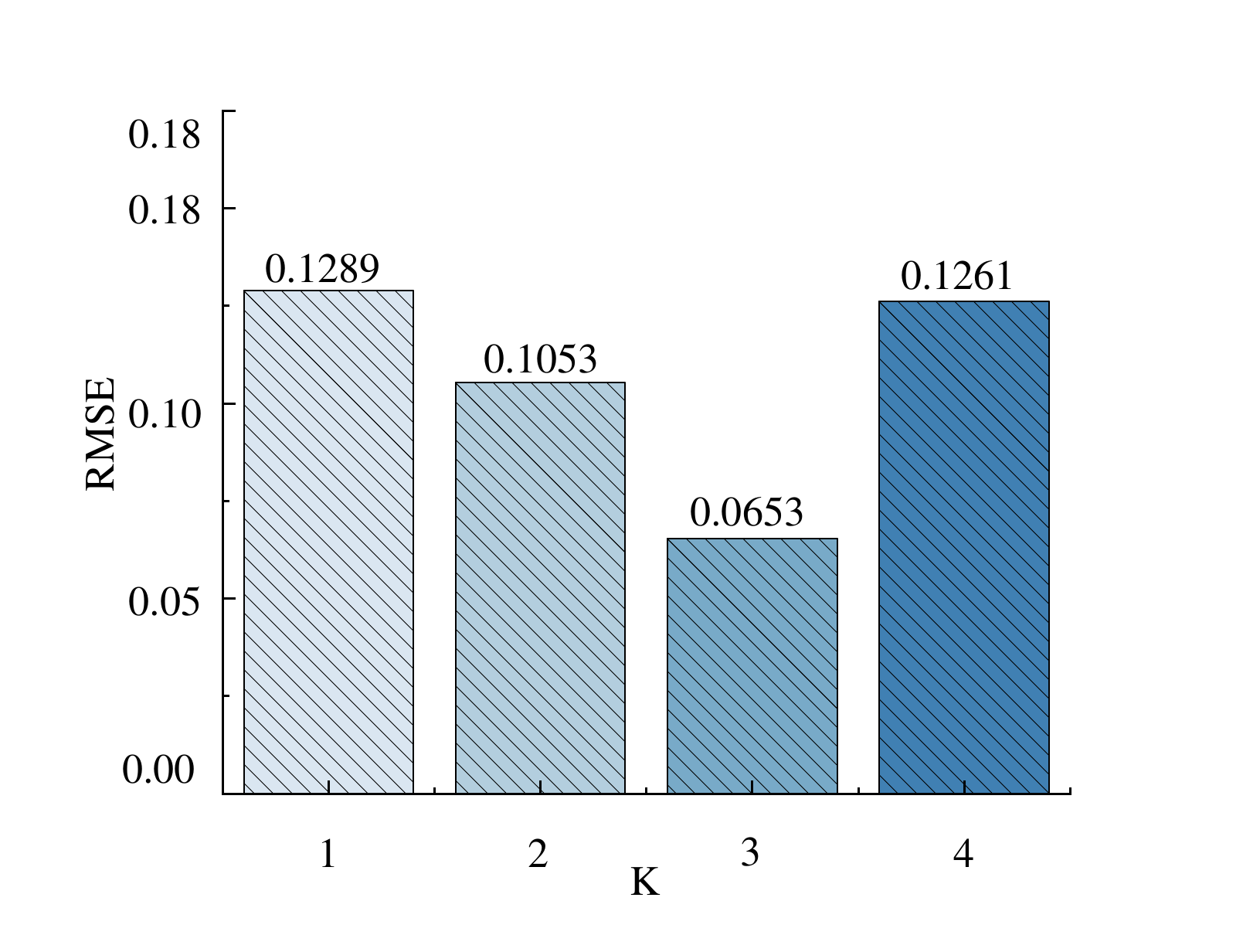} 
	}
  \vspace{-0.2cm}
	\caption{Impact of order $K$ on the Samson dataset, where (a) SAD and (b) RMSE.}\label{figK}
\end{figure}

Regarding the propagation depth $K$, increasing $K$ does not necessarily lead to consistent performance gains. Although a larger $K$ enables the model to exploit information from more distant neighborhoods, it may also introduce issues such as over-smoothing. The learnable weight $\alpha$ alleviates this trade-off to some extent, since the model can assign near-zero weights to ineffective higher-order propagation, thereby avoiding their negative effects. We further provide an ablation study with $K \in \{1,2,3,4\} $ in Fig. \ref{figK}. The results show that performance typically improves with a moderate $K$ and then saturates. Accordingly, we select $K =3$ as the best trade-off between accuracy and computational complexity.

\section{Conclusion}\label{Conclusion}
This letter introduces a novel transformer-guided content-adaptive graph unmixing method. Technically, it leverages the transformer to capture global dependencies, while incorporating content-adaptive graph construction and multi-order propagation fusion to enhance local consistency and boundary precision without sacrificing global priors. Moreover, the introduction of a graph residual mechanism strengthens representation capability while maintaining training stability. Experimental results validate the superior performance of our proposed method on both simulated and real HSI datasets. Future work will develop lightweight network architectures to reduce complexity while maintaining accuracy.

\bibliographystyle{elsarticle-num}
\bibliography{IEEEabrv,mybibfile}

\end{document}